%% file: 00_main.tex
\documentclass[10pt,twocolumn,letterpaper]{article}

\usepackage{cvpr}              

\usepackage{comment}

\usepackage{graphicx}
\usepackage{times}
\usepackage{epsfig}
\usepackage{multirow}
\usepackage{listings}
\usepackage{xcolor}
\definecolor{codegreen}{rgb}{0,0.6,0}
\definecolor{codegray}{rgb}{0.5,0.5,0.5}
\definecolor{codepurple}{rgb}{0.58,0,0.82}
\definecolor{backcolour}{rgb}{0.95,0.95,0.92}

\lstdefinestyle{mystyle}{
    backgroundcolor=\color{backcolour},   
    commentstyle=\color{codegreen},
    keywordstyle=\color{magenta},
    numberstyle=\tiny\color{codegray},
    stringstyle=\color{codepurple},
    basicstyle=\ttfamily\scriptsize,
    breakatwhitespace=false,         
    breaklines=true,                 
    captionpos=b,                    
    keepspaces=true,                 
    showspaces=false,                
    showstringspaces=false,
    showtabs=false,                  
    tabsize=2
}

\usepackage{amsmath}
\usepackage{amssymb}
\usepackage{url}
\usepackage{multirow}
\usepackage{xspace}
\usepackage{color}
\usepackage{blindtext} 
\usepackage{epigraph} 
\usepackage{xspace}

\usepackage{amssymb}
\usepackage{pifont}

\usepackage{nicefrac}       
\usepackage{microtype}      

\usepackage{wrapfig}
\usepackage{enumitem}
\usepackage{booktabs}
\usepackage{floatrow}
\usepackage{caption}
\usepackage{listings}
\usepackage{enumitem}
\usepackage{wrapfig}
\usepackage{lipsum}
\usepackage{graphics}
\usepackage{listings}
\usepackage{xcolor}
\usepackage{algorithm}
\usepackage{float}
\RequirePackage{algorithmic}
\DeclareMathOperator*{\argmax}{arg\,max}

\newfloatcommand{capbtabbox}{table}[][\FBwidth]


\newcommand{\katef}[1]{{\color{magenta} }}
\input{macros}

\input{preamble}

\definecolor{cvprblue}{rgb}{0.21,0.49,0.74}
\usepackage[pagebackref,breaklinks,colorlinks,citecolor=cvprblue]{hyperref}

\def\paperID{10149} 
\def\confName{CVPR}
\def\confYear{2024}

\title{Diffusion-ES: Gradient-free Planning with Diffusion for Autonomous Driving and Zero-Shot Instruction Following}

\author{
Brian Yang
\qquad
Huangyuan Su
\qquad 
Nikolaos Gkanatsios
\qquad 
Tsung-Wei Ke \\
\qquad 
Ayush Jain
\qquad 
Jeff Schneider
\qquad
Katerina Fragkiadaki\\
Carnegie Mellon University
}

\begin{document}
\maketitle
\begin{figure*}[h!]
  \centering
    \includegraphics[width=0.97\linewidth]{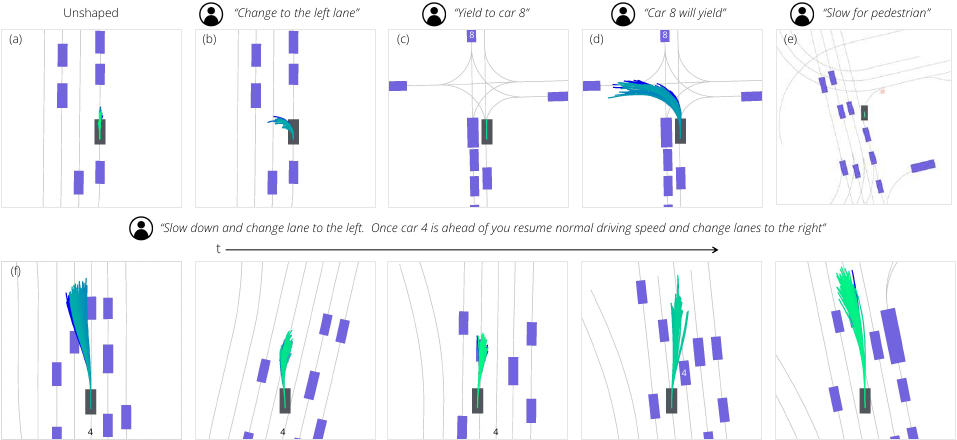}
  \caption{\textbf{\model{} is a  test-time trajectory optimization method for arbitrary reward functions that combines generative trajectory models and sampling-based search.}  (a) Trajectories generated by \model{} by optimizing a general driving reward function that encourages  following lanes, avoiding collisions, and respecting traffic signs. \model{}  achieves state-of-the-art driving performance  in nuPlan \cite{caesar2022nuplan}. (b-f) \model{}  follows driving instructions in natural language through test-time optimization of language-shaped reward functions, without any additional training.  We  prompt  LLMs to map  language instructions to programs that shape the  driving reward function, which we then optimize with   \model{}. }
  \label{fig:teaser}
\end{figure*}

\input{1_abstract}    

\input{2_intro}

\input{3_related}
\input{4_method}
\input{5_experiments}
\input{6_conclusion}

\input{7_acknowledgements}
\bibliographystyle{ieeenat_fullname}
\bibliography{bibs/refs,bibs/driving,bibs/darpa,bibs/ebm,bibs/chain}
\input{8_appendix}


\end{document}

%% file: macros.tex
\newcommand{\bzero}{\mathbf{0}}
\newcommand{\bone}{\mathbf{1}}
\newcommand{\cond}{\textbf{c}}


%% file: preamble.tex
\newcommand{\model}{Diffusion-ES}


%% file: 1_abstract.tex
\begin{abstract} 

 Diffusion models excel at modeling  complex and multimodal trajectory distributions for decision-making and control.  
 Reward-gradient guided denoising has been recently proposed to generate  trajectories that maximize both a differentiable reward function and the likelihood under the data distribution
 captured by a diffusion model.
Reward-gradient guided denoising requires a differentiable reward function fitted to both clean and noised samples, limiting its applicability as a general trajectory optimizer.  
In this paper, we propose \model{}, a method that combines gradient-free optimization with trajectory denoising to optimize black-box non-differentiable objectives while staying in the data manifold. 
 \model{} samples trajectories during evolutionary search from a diffusion model and scores them using a black-box reward function. 
 It mutates high-scoring trajectories using a truncated diffusion process that applies a small number of noising and denoising steps, allowing for much more efficient exploration of the solution space. 
 We show that \model{} achieves state-of-the-art performance on nuPlan, an established closed-loop planning benchmark for autonomous driving.
 \model{} outperforms existing sampling-based planners, reactive deterministic or diffusion-based  policies, and reward-gradient guidance.
 Additionally, we show that unlike prior guidance methods, our method can optimize non-differentiable language-shaped reward functions generated by few-shot LLM prompting.
 When guided by a human teacher that issues instructions to follow, our method can generate novel, highly complex behaviors, such as aggressive lane weaving, which are not present in the training data.
 This allows us to solve the hardest nuPlan scenarios which are beyond the capabilities of existing trajectory optimization methods and driving policies.
\footnote{Project page: \url{diffusion-es.github.io}}
\end{abstract}

%% file: 2_intro.tex

\section{Introduction}

  
Diffusion models have   shown to excel at modeling highly complex and multimodal trajectory distributions for decision-making and control \cite{janner2022diffuser, Ajay2022IsCG}.
Reward-gradient guidance  \cite{janner2022diffuser,urain2022se,Liang2023AdaptDiffuserDM} has been used to test-time optimize differentiable reward functions by alternating between denoising diffusion steps and backpropagating reward gradients to the noised trajectory. In this way, sampled trajectories are pushed towards the trajectory data manifold  while also maximizing the reward function at hand \cite{janner2022diffuser}. This decoupling of the reward function  from  trajectory diffusion permits a single trajectory diffusion  model to be used for maximizing a variety of  reward functions at test time. Reward-gradient guidance  requires the reward function to be differentiable and fitted in both noisy and clean trajectories, which usually requires re-training. This limits its applicability as a general solver for trajectory optimization.

We propose \model{}, a reward-guided denoising  method  for 
 optimization of non-differentiable, black-box objectives that samples and mutates trajectories using a diffusion model, guided by a reward function that operates only on the clean, final, denoised samples. 
Naively combining diffusion with sampling-based optimization does not work: sampling-based optimizers, like CEM \cite{rubinstein1997optimization} or MPPI \cite{williams2015model}, typically require a large population of samples across multiple  iterations of selection and mutation  to converge to good solutions, which, when combined with the computational cost of denoising inference, results in a prohibitively slow search process. 
In \model{}, \textit{high-scoring trajectories are mutated using a truncated diffusion-denoising process}, by adding a small amount of noise  and denoising them back, as shown in Figure \ref{fig:architecture} \textit{right}. The amount of added noise is progressively decreased across search iterations making \model{} computationally viable. 

The trajectory diffusion model used for test-time optimization in \model{} can in principle condition on any scene-relevant information to narrow the sampling to a distribution of scene-relevant trajectories. 
In fact, the amount of conditioning information controls a continuum between train-then-test learning  and test-time planning, and a corresponding trade-off  between inference speed and out-of-distribution (OOD) generalization: \textbf{1.} \textit{The more conditioning information, the narrower the distribution to draw trajectory samples from, the faster the search.} In the extreme, no test-time reward optimization is used and our diffusion model operates as a reactive policy at test time. 
Indeed, many methods train diffusion policies \cite{chi2023diffusion} as \textit{conditional} diffusion trajectory prediction models 
using rewards as conditioning information to the trajectory diffusion model \cite{Ajay2022IsCG} or finetune a diffusion policy with reinforcement learning \cite{wang2022diffusion,lu2023contrastive} or imitation learning \cite{pearce2023imitating}. While these methods can handle black-box reward functions or good behaviours to imitate, we show that using them as is or with reward-guidance at test time often under-performs reward-guided denoising of an \textit{unconditional} diffusion model, which completely decouples trajectory and reward modelling.
\textbf{2.} \textit{The less conditioning information, the wider the distribution to draw trajectory samples from, the slower the search, but the better the  generalization to OOD tasks and scenarios} that require novel pairings of trajectories and scene contexts, not present in the training data. 
Indeed, this is the premise of test-time planning over train-then-test  learning: test-time optimization of a composition of energy functions \cite{gkanatsios2023energybased,du2023reduce}, here, the energy of the trajectory data distribution and the energy of arbitrary reward functions, should be able to synthesize novel behaviours not seen at training time.

We show how \model{}, when combined with an unconditional diffusion model over trajectories, can achieve state-of-the-art planning performance purely through diffusion-guided black-box reward maximization.
Our approach is evaluated on nuPlan \cite{caesar2022nuplan}, an established driving benchmark built on real-driving logs and estimated ground-truth perception.
We achieve state-of-the-art performance for closed-loop driving, matching the performance of the previous SOTA, PDM-Closed, a sampling-based planner tailored to the nuPlan benchmark \cite{dauner2023parting}, as well as reactive driving policies \cite{scheel2022urban,plant}, deterministic or diffusion-based.
Moreover, we illustrate the flexibility of \model{} by test-time optimizing language-shaped reward functions generated using few-shot LLM prompting.
Using language instructions, we can solve the most challenging nuPlan scenarios, as well as synthesize entirely novel driving behaviors.
We then test our model and baselines in their ability to optimize the generated reward functions to elicit the desired behaviours. 
Qualitative examples of behaviors generated by instruction following using our method can be found in Figure \ref{fig:teaser}.
We show \model{} dramatically outperforms PDM-Closed
, other sampling-based planners, as well as ablative versions of \model{} that either condition the diffusion model on the surrounding scene,  
or do not use any guidance at all.

In summary, our contributions are as follows: 
\begin{enumerate}
\item We introduce \model{},  a trajectory optimization method for optimizing  black-box objectives that uses a trajectory diffusion model for sampling and mutating trajectory proposals during sampling-based search. We show \model{} matches the  SOTA performance of engineered planners  in closed-loop driving in nuPlan, and much outperforms them when optimizing more complex reward functions that require flexible driving behaviour, beyond lane following. 
 To the best of our knowledge this is the first work to combine evolutionary search with diffusion models. 
 

 \item We show that \model{} can be used to follow language instructions and steer the closed-loop driving behaviour of an autonomous vehicle by optimizing the LLM-shaped reward functions, \textit{without any training data of language and actions}. We showed that such instruction following  can solve the most challenging driving scenarios in nuPlan. 
 

\item We show extensive ablations of our model with varying amount of conditioning information which  clearly reveals the trade off between inference speed and out-of-distribution (OOD) generalization in driving. 

\end{enumerate}
We believe \model{} will be useful to the community as a general trajectory optimizer with applicability beyond driving. Our code and models will be publicly available upon publication to aid reproducibility in the project webpage: \url{diffusion-es.github.io}.


%% file: 3_related.tex
\section{Related work}

\noindent \textbf{Diffusion models for decision-making and trajectory optimization} Diffusion models \cite{SohlDickstein2015DeepUL,ho2020denoising,song2020denoising,nichol2021improved} learn to approximate the data distribution through an iterative denoising process and have shown impressive results on image generation \cite{Dhariwal2021DiffusionMB,Saharia2022PhotorealisticTD,Rombach2021HighResolutionIS,Ramesh2022HierarchicalTI}. 
They have been used for imitation learning for manipulation tasks \cite{chi2023diffusion,reuss2023goal,pearce2023imitating,XianChainedDiffuserUT}, for controllable vehicle motion generation \cite{Zhong2022GuidedCD,Jiang_2023_CVPR,Carvalho2023MotionPD} and for video generation of manipulation tasks \cite{Du2023LearningUP,Yang2023LearningIR}. 
Works of \cite{controllabletraffic,MotionDiffuser} use  diffusion models to forecast offline vehicle trajectories. 
To the best of our knowledge, this is the first work to use diffusion models in closed-loop driving. 

\noindent \textbf{Learning versus planning for autonomous driving}
Learning to drive from imitating  driving demonstrations is prevalent in the research and development of autonomous vehicles \cite{DBLP:journals/corr/abs-1710-02410,gaildriving,chauffernet,plant,roach,alvinn,endtoendNVIDIA,LBC,LAV,transfuser}.  
Many preeminent imitation methods assume the underlying action distribution is unimodal, which is problematic when training from multimodal expert demonstrations. 
Objectives and architectures that can better handle multimodal trajectory prediction have been proposed \cite{wayformer,shafiullah2022behavior,Kim_2023,multiplefutures, fragkiadaki2017motion,lookout}.
We show that diffusion models are well-suited for driving and can be used to synthesize rich complex behaviors from multimodal demonstrations.

On the other hand, conventional autonomy stacks do not rely on learning at all for decision making, and rather rely on optimizing manually engineered cost functions online \cite{fan2018baidu,montremerlo2008stanford,ziegler2014trajectory}.
Recently, PDM-Closed \cite{dauner2023parting} achieved state-of-the-art performance on the nuPlan driving benchmark by purely relying on test-time planning and heuristics for selecting trajectory proposals.
Other prior works aim to incorporate the benefits of offline learning for test-time planning by performing sampling-based planning over learned cost maps \cite{zeng2019end} or doing gradient-based optimization over learned dynamics models \cite{rhinehart2021contingencies}.
We extend this line of work by showing how diffusion-based generative models can be combined with sampling-based planning.

\noindent \textbf{Language-conditioned policies for autonomous driving} Recently there has been significant progress made towards language-conditioned policies for driving.
GAIA-1 \cite{hu2023gaia} is a generative world model capable of multimodal video generation that leverages video, language and actions to synthesize driving scenarios which can comply with given language instructions.
However, GAIA-1 does not execute any actual control inputs.

LLMs trained from Internet-scale text have shown impressive zero-shot reasoning capabilities for a variety of downstream language tasks when prompted appropriately, without any weight fine-tuning \cite{DBLP:journals/corr/abs-2201-11903,DBLP:journals/corr/abs-2107-13586,brown2020language}. 
Recent works have shown that LLMs can be prompted to map language instructions to language subgoals \cite{DBLP:journals/corr/abs-2012-07277,DBLP:journals/corr/abs-1909-13072,huang2022language,https://doi.org/10.48550/arxiv.2207.05608}  action programs \cite{liang2022code,surís2023vipergpt,Gupta_2023_CVPR} or cost maps \cite{huang2023voxposer} with appropriate plan-like or program-like prompts.  
Our work follows few-shot prompting of LLMs to shape  driving reward functions.  We extend previous methods by using Python generators to produce reward functions which maintain internal state across calls.  
Works of \cite{mao2023gpt,wen2023road,xu2023drivegpt4} use LLMs to predict low-level control signals given high-level scene descriptions and language instructions.
However, none of these approaches evaluate their performance on closed-loop driving, which is significantly more difficult than open-loop trajectory forecasting \cite{dauner2023parting}.
\cite{sha2023languagempc} is similar to us, but only considers a highly simplified driving setup and does not report results on a standardized benchmark with strong baselines.

%% file: 4_method.tex
\begin{figure*}[h!]
  \centering
    \includegraphics[width=0.975\linewidth]{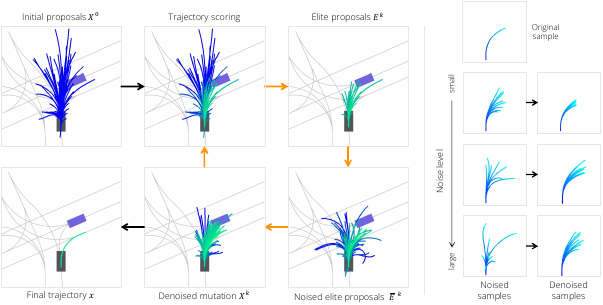}
  \caption{\textbf{Overview for \model{}.} \textit{Left:} Generative evolutionary search with \model{}. Trajectories are colored by rewards (\textcolor{blue}{blue} is low, \textcolor{green}{green} is high). \textit{Right:} We visualize the mutations for varying noise levels. Color denotes timestep along trajectory. While noise perturbations alone can lead to unrealistic trajectories, denoising helps project samples back onto the trajectory data manifold.
  }
  \label{fig:architecture}
\end{figure*}

\section{Method}  \label{sec:method}


\subsection{Background} \label{sec:background}

\noindent \textbf{Diffusion models} 
A diffusion model captures the probability distribution $p(x)$ through the inversion of a forward diffusion process, that gradually adds Gaussian noise to the intermediate distribution of a initial sample $x$.  The amounts of added noise depend on a predefined variance schedule ${\beta_t \in (0,1)}_{t=1}^{T}$, where $T$ denotes the total number of diffusion timesteps.  At diffusion timestep $t$, the forward diffusion process adds noise into $x$ using the formula $x_t = \sqrt{\bar{\alpha}_t} x + \sqrt{1-\bar{\alpha}_t} \epsilon$, where $\epsilon \sim \mathcal{N}(\bzero,\bone)$ is a sample from a Gaussian distribution with the same dimensionality as $x$. Here, $\alpha_t = 1- \beta_t$, and $\bar{\alpha}t = \prod{i=1}^t \alpha_i$.  For denoising, a neural network $\hat \epsilon = \epsilon_{\theta}(x_t;t)$ takes input as the noisy sample $x_t$ and the diffusion timestep $t$, and learns to predict the added noise $\epsilon$.  
To generate a sample from the learned distribution $p_\theta(x)$, we start by drawing a sample from the prior distribution $x_T \sim \mathcal{N}(\mathbf{0},\mathbf{1})$ and iteratively denoise this sample $T$ times with $\epsilon_\theta$.  The application of $\epsilon_\theta$ depends on a specified sampling schedule~\cite{ho2020denoising,song2020denoising}, which terminates with $x_0$ sampled from $p_\theta(x)$.
Diffusion models can be easily extended to model $p(x|\cond)$, where $\cond$ is some conditioning signal, such as the expected future rewards 
, by adding an additional input to the denoising neural network $\epsilon_{\theta}$.

\noindent \textbf{Evolutionary strategies}
Evolutionary strategies (ES) are a family of population-based \textit{gradient-free} optimization algorithms which can maximize arbitrary black-box reward functions $R(x) : \mathbb{R}^{d} \rightarrow \mathbb{R}$ without any  training, where $x$ is the variable we optimize over. 
ES iteratively update a search distribution $q(x)$ to maximize expected rewards $\mathbb{E}_{x\sim p_{\theta}(x)}[R(x)]$.
While distribution-based ES approaches such as CEM and CMA-ES represent $q(x)$ explicitly (often as a unimodal Gaussian), we can also represent $q(x)$ non-parametrically as a set of high-performing solutions without making strong assumptions about the functional form of $q$. 

\subsection{\model{}} \label{sec:ourmodel}
\model{} is a  trajectory optimization method  that leverages gradient-free evolutionary search to perform  reward-guided sampling from trained diffusion models, for any black-box reward function $R(x)$.  
Specifically, we use a trained  diffusion model  $\epsilon_\theta$ to initialize the sample population and we use a truncated diffuse-denoise process to mutate samples while staying in the data manifold. 
The control flow of \model{} is shown in Figure \ref{fig:architecture} (left) and in Algorithm \ref{alg:diff-tta}. 

\noindent \textbf{Initializing the population with diffusion sampling}
We begin by sampling an initial population $X^{0}$ of $M$ trajectory samples using our diffusion model:
\begin{align}
    X^{0} = \{ x_{i}\}_{i=1}^{M} \sim p_{\theta}(x),
\end{align}
where $X^{k}$ is the population at iteration $k$.
This involves a complete pass through the reverse diffusion process. Given we use an unconditional diffusion model, these samples are scene agnostic and can always be used without re-sampling them at each timestep. 
We can also modify the initial population  by including samples generated by other approaches or mixing in solutions from the previous timestep to warm-start our optimization. 

\noindent \textbf{Sample scoring}
At each iteration $k$, we score the samples  in our population $\{ R(x_{i}) | x_{i} \in X^{k}\}_{i=1}^{M}$. 
Note that our population consists of ``clean" samples so we do not need a reward function which can handle "noisy" samples, which gives us significant flexibility compared to guidance methods that perform classifier-based guidance.

\noindent \textbf{Selection}
We use rewards to decide which samples we should select to propagate to the next iteration.
Similar to MPPI \cite{williams2015model}, we resample $X^{k+1}$ as follows:
\begin{align}
    q(x) &= \frac{\exp (\tau R(x))}{\sum_{i=1}^{M} \exp (\tau R(x_i))} \\
    E^{k+1} &= \{ x_{i} \overset{\mathrm{iid}}{\sim} q(x)\}_{i=1}^{M},
\end{align}
where $E^{k+1}$ represents our elite set which is kept from iteration $k$, and $\tau$ is a tunable temperature parameter controlling the sharpness of $q$.

\noindent \textbf{Mutation using truncated diffusion-denoising}
We apply randomized mutations to $E^{k+1}$ for  exploration. 
Prior evolutionary search methods resort to naive Gaussian perturbations which do not exploit any prior knowledge about the data manifold.
\textbf{Our key insight is to leverage a truncated diffusion-denoising process to mutate  trajectories in a way the resulting mutations are part of the  data manifold.} 
We can run the first $t$ steps of the forward diffusion process to get noised elite samples $\bar{E}^{k+1}$:
\begin{align}
    \bar{E}^{k+1} = \{\sqrt{\bar{\alpha}_N} x + \sqrt{1-\bar{\alpha}_N} \epsilon | x \in E^{k+1} \},
\end{align}
where $\epsilon \sim \mathcal{N}(\bzero,\bone) $.
Then we can run the last $t$ steps of the reverse diffusion process to denoise the samples again, giving us clean samples $X^{k+1}$:
\begin{align}
    X^{k+1} = \{x \sim p_\theta(x | \bar{x}) | \bar{x} \in \bar{E}^{k+1}\}. 
\end{align} 
In practice, the number of timesteps $t$ of the truncated diffusion process is a tunable time-dependent hyperparameter $t_k$ which controls the mutation strength at each iteration $k$.
This is visualized in Figure \ref{fig:architecture} (right).
We find that linearly decaying the number of mutation diffusion steps $t_k$ from 5 to 1 over 20 search steps works best in our experiments.


\begin{algorithm}[H]
   \caption{$\texttt{\model{}}$}
   \label{alg:diff-tta}
\begin{algorithmic}[1]
    \STATE {\bfseries Input:} Diffusion model $p_\theta$, reward function $R$, search steps $K$, population size $M$, number of noising steps $N$, variance schedule $\bar{\alpha}=\{\bar{\alpha}_i\}_{i=1}^N$.
    \STATE {\bfseries Initial proposals:} $X^0 \leftarrow \{x_i \sim p_\theta (x)\}_{i=1}^M$
    \FOR{search step $k \in (1, \dots, K)$}
        \STATE Score proposals $\{R(x_i) | x_i \in X^{k-1}\}_{i=1}^M$
        \STATE Compute distribution $q (x)=\frac{\exp(\tau R(x))}{\sum_{i=1}^{M} \exp(\tau R(x_i))}$
        \STATE Sample elites from $X^{k-1}$: $E^{k} \leftarrow \{x_i \overset{\mathrm{iid}}{\sim} q(x)\}_{i=1}^M$
        \STATE Renoise elites
        $\bar{E}^{k} \leftarrow \{\sqrt{\bar{\alpha}_N} x + \sqrt{1-\bar{\alpha}_N} \epsilon | x \in E^k , \epsilon \sim \mathcal{N}(\bzero,\bone) \}$
        \STATE Denoise elites $X^{k} \leftarrow \{x \sim p_\theta(x | \bar{x}) | \bar{x} \in \bar{E}^k\}$
    \ENDFOR
    \RETURN output $x = \argmax_{x \in X^K} R(x)$
\end{algorithmic}
\end{algorithm}

\subsection{Mapping language instructions to reward functions with LLM prompting} 

\begin{figure*}[h!]
  \centering
    \includegraphics[width=0.975\linewidth]{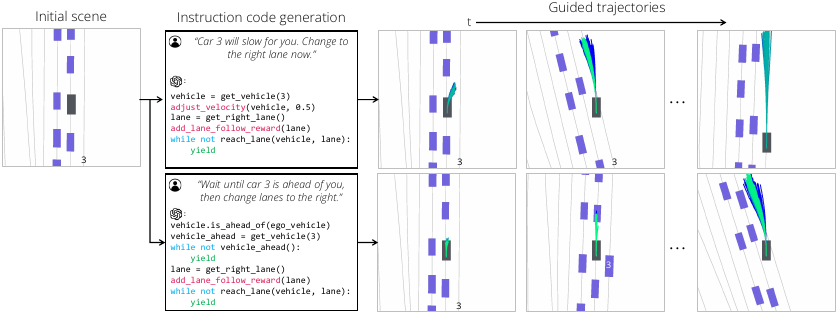}
  \caption{\textbf{Shaping  reward functions with language instructions using LLM prompting and  optimizing them with  \model{}.} 
  For the same initial scene, we show two distinct language instructions, along with their generated reward shaping code and generated  trajectories from \model{}. 
  }
  \label{fig:lang}
\end{figure*}

\label{sec:lang}
To follow driving instructions given in natural language, we map them to black-box reward functions which we optimize with \model{}.  
We adopt a similar approach to \cite{yu2023language,gen2sim} which uses LLMs to synthesize reward functions from language instructions. 
Reward functions are composable and allow us to seamlessly combine language guidance with other constraints.
This is crucial in driving where we  constantly optimize many different objectives at once (e.g., safety, driver comfort, route adherence).

Similar to prior work, we expose a Python API which can be used to extract information about entities in  the road scene. 
Since many of the basic reward signals in driving do not change from scenario to scenario (e.g., collision avoidance or drivable area compliance), we allow the LLM to write \textit{reward shaping} code which modifies the behavior of the base reward function, as opposed to generating everything from scratch. 
Reward shaping can add auxiliary reward terms (e.g., a dense lane-reaching reward) or re-weight existing reward terms. We show generated code examples  in Figure \ref{fig:lang}. 

Our goal is to handle general and complex language instructions with temporal dependencies, such as \textit{``Change lanes to the left, then pass the car on the right, then take the exit"}. 
Previous works \cite{yu2023language,huang2023voxposer} produce a stationary reward function, i.e., one which is fixed during planning. 
This can make it challenging to express sequential plans solely through rewards.   
We find that a much more natural and succinct way of capturing these plans in code is through the use of \textit{generator functions} which retain internal state between calls. 
%
All our prompts and code examples can be found  in the supplementary file. In Section \ref{sec:exps}, we show how optimizing these language-shaped reward functions can synthesize rich and complex driving behaviors that comply with the language instructions. 

%% file: 5_experiments.tex
\section{Experiments}
\label{sec:exps}
We first evaluate \model{} on closed-loop driving in nuPlan   \cite{caesar2022nuplan}, an established  benchmark  that uses estimated perception for vehicles, pedestrians, lanes and traffic signs. Our model and baselines are evaluated on their ability to drive safely and efficiently while having access to  close-to-ground-truth perception output provided by the dataset. 
We also consider a suite of driving instruction following tasks. 
We map instructions to shaped reward functions with LLM prompting, and evaluate \model{} and baselines in their ability to optimize the generated reward functions and accurately follow the instructions.  
Our experiments aim to answer the following questions:
\begin{enumerate}
\item How does \model{} compare to existing  sampling-based planners and  reward-gradient guidance for trajectory optimization? 
\item How does \model{} compare to SOTA reactive driving policies that directly map environments' state to  vehicle trajectories? 
\item Can the hardest nuPlan driving scenarios be solved by assuming  access to a human teacher giving instructions in natural language, without any additional training data?
\item Does scene conditioning for the diffusion model benefit \model{}?
\end{enumerate}

\subsection{Closed-loop driving}
\label{sec:closedloop}
We evaluate our model on the nuPlan Val14 planning benchmark \cite{dauner2023parting}. 
We specifically consider the reactive agent track of the nuPlan benchmark since it is the most difficult and realistic of the evaluation settings in nuPlan.

\paragraph{Model setup and hyperparameters}
For \model{}, we train a diffusion model over ego-vehicle trajectories consisting of 2D poses $(x,y,\theta)$ predicted 8 seconds into the future at 2Hz, leading to an overall action dimension of 48.
Unlike prior work \cite{janner2022diffuser}, we model the distribution over actions only rather than modeling states and actions.
We use a population size of $M=128$ in our experiments. Our diffusion model is trained with $T=100$ denoising steps.

\paragraph{Reward function} 
We adopt a modified version of the scoring function used in PDM-Closed \cite{dauner2023parting} as our reward function.
To compute rewards, we convert our predicted trajectory to low-level control inputs using an LQR tracker. These control inputs are fed into a kinematic bicycle model \cite{polack2017kinematic} which propagates the dynamics of the ego-vehicle. We follow \cite{dauner2023parting} and forecast the motion of other agents by assuming constant velocity.
These simulated rollouts are then scored following the nuPlan benchmark evaluation metrics.
We also add auxiliary reward terms to penalize proximity to the leading agent and enforce speed limits.

Note that this reward function is \textit{not differentiable} due to the tracker and the use of non-differentiable heuristics for assessing traffic violations.
Additionally, training a model to regress rewards is challenging since the nuPlan dataset contains no instances of serious traffic infractions.


\paragraph{Evaluation metric} We report our results using \textbf{driving score}, which aggregates multiple planning metrics related to traffic rule compliance, safety, route progress, and rider comfort. This is the standard evaluation metric used in nuPlan.

\paragraph{Baselines}
We consider the following baselines:
\begin{itemize}
    \item \textit{UrbanDriverOL} \cite{Scheel2021UrbanDL}, a deterministic transformer policy trained with behaviour cloning and augmentations.
    Unlike in \cite{Scheel2021UrbanDL}, the nuPlan implementation does not perform closed-loop training.
    \item \textit{PlanCNN} \cite{plant} a deterministic imitation policy which encodes a rasterized BEV map using a CNN backbone.
    \item \textit{IDM} \cite{helbing1998generalized}: a heuristic rule-based planner which adjusts its speed to maintain a safe distance to the leading vehicle. It is also used to control the behaviour of agents in nuPlan.
    \item \textit{PDM-Closed} \cite{dauner2023parting}: an MPC-based planner which generates path proposals using lane centerlines, and rolls out trajectories similarly to us. Instead of iteratively optimizing rewards, PDM-Closed simply executes the highest-performing proposal after one round of scoring. It is the current state-of-the-art on the nuPlan Val14 benchmark. 
    \item \textit{Diffusion Policy}: a diffusion model we consider that conditions on scene features to predict  a vehicle trajectory directly.
    We encode scene features using the transformer feature backbone from Urban Driver \cite{scheel2022urban}. We train it with imitation learning and augmentations. This is similar to the unconditional trajectory model in \model{} with additional conditioning on scene features. 
\end{itemize}

\begin{table}
  \centering
  \begin{tabular}{l|lc}
    \toprule
    & Method & Driving Score ($\uparrow$) \\
    \midrule
    \parbox[t]{5mm}{\multirow{3}{*}{\rotatebox[origin=c]{90}{\parbox[c]{1cm}{\centering \footnotesize Train-then-test}}}} & 
    UrbanDriverOL \cite{scheel2022urban} & 65\\
    & PlanCNN \cite{plant} & 72\\
    & Diffusion policy & 50 \\
    \midrule
    \parbox[t]{2mm}{\multirow{3}{*}{\rotatebox[origin=c]{90}{\parbox[c]{1.2cm}{\centering \footnotesize Test-time optimize}}}} & IDM \cite{rubinstein1997optimization}  & 77\\
    & PDM-Closed \cite{dauner2023parting} & \textbf{92}\\
    & Diffusion-ES (ours) & \textbf{92} \\
    \bottomrule
  \end{tabular}
  \caption{\textbf{Closed-loop driving results in the val14 split of \cite{dauner2023parting}.} Our model matches the performance of previous non-learning based planners and significantly outperforms all other models.}
  \label{tab:example}
\label{tab:val14}
\end{table}

We show  quantitative results in Table \ref{tab:val14}. We draw the following conclusions:

\noindent \textbf{1. \model{} matches the prior state-of-the-art, PDM-Closed} and substantially outperforms all other baselines. 
PDM-Closed is a sampling-based planner that relies on domain-specific heuristics to generate trajectory proposals whereas \model{} learns these proposals from data. Both use a similar reward function and dynamics model for the agents in the scene.

\noindent \textbf{2. There is a large gap in performance between reactive neural policies and test-time planners},  also pointed  out in recent work \cite{dauner2023parting}. 
We hypothesize that this is because compared to other control benchmarks, nuPlan has a much richer observation space as scenes are densely populated by dynamic actors, many of which are irrelevant to the ego-agent.
This can make it challenging for learning-based methods to generalize, which motivates the need of  test-time optimization. 

\noindent \textbf{3. \model{}  substantially outperforms  diffusion policy. } 
Qualitatively, our  diffusion policy has a tendency to randomly change lanes, which  causes the ego-vehicle to reach out-of-distribution scenarios faster. 
\model{} leverages the expressiveness of generative modeling while using test-time optimization to improve generalization.

\subsection{Language instruction following}
One drawback of the nuPlan driving benchmark is that encourages highly conservative driving behaviors. 
For example, PDM-Closed \cite{dauner2023parting} holds the state-of-the-art in the nuPlan Val14 benchmark while being unable to change lanes, since its path proposals only consider the lane the ego-vehicle is currently on. However, lane changing is not necessary for good driving performance in the current nuPlan benchmark.

To evaluate  \model{} and baselines in their ability to 
optimize arbitrary reward functions, we consider eight language instruction following tasks, 
each taken from an existing driving log in the nuPlan benchmark. 
In each task, the language instruction  requires the ego-vehicle to perform a specific driving maneuver that solves a challenging driving scenario.
In most scenarios there will be no examples of the instructed behavior anywhere in nuPlan. 
For instance, the lane weaving task requires the ego-vehicle to aggressively change multiple lanes in dense urban traffic.
Task descriptions, language instructions, and prompts are provided in the appendix. 
We use the method described in \ref{sec:lang} to generate executable Python code given a language instruction that adapts the initial reward function of Section \ref{sec:closedloop}, giving us a language-shaped reward function for each scenario. Our model and baselines  will optimize the same language-shaped reward function.

\paragraph{Evaluation metric}
We evaluate our model and baselines using 
\textbf{task success rate}, which measures how frequently the agent was able to successfully complete the designated task. 
To increase the difficulty of the tasks, we randomize the behavior of other vehicle agents by adding noise to their IDM parameters at sporadic intervals during each episode.
All scores reported are averaged across ten random seeds.

\paragraph{Baselines}
We compare to the following baselines:
\begin{itemize}
    \item \textit{PDM-Closed}: this is adapted to this setting by using our language-shaped reward function in place of the original reward function.
    \item \textit{PDM-Closed-Multilane}: a modified variant of PDM-Closed which considers a wider range of laterally offset paths, allowing for lane changes.
    \item \textit{Conditional \model{}}: \model{} that uses a conditional diffusion model instead of an  unconditional one.
\end{itemize}

\vspace{-0.1in}
\begin{figure}[h!]
  \centering
    \includegraphics[width=0.975\linewidth]{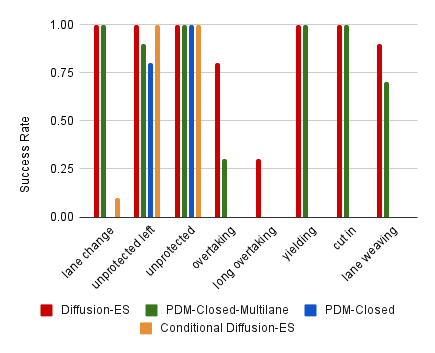}
  \caption{\textbf{Following driving instructions.} \model{} outperforms all baselines in optimizing complex language-shaped reward functions.  
  }
  \label{fig:controllability_success}
\end{figure}

\noindent Figure \ref{fig:controllability_success} shows the success rates on the controllability tasks. We draw the following conclusions:

\noindent \textbf{1. \model{}  outperforms all baselines}. Although PDM-Closed-Multilane has substantially improved performance compared to PDM-Closed due to more diverse trajectory proposals, it is still weaker than Diffusion-ES on 4 out of 8 tasks. This highlights the weakness of relying on handcrafted rules for proposal generation.
\noindent \textbf{2. Diffusion-ES performs significantly worse with a conditional diffusion model.}
The conditional diffusion model is much harder to guide since the scene context causes fewer samples to be in-distribution.

\subsection{Lane following}
To compare \model{} against reward-gradient guidance, we consider a simplified lane following task with a differentiable reward function, which consists of two terms:
one penalizing lateral deviation from the lane (\textbf{lane error}), and one penalizing deviation from a target speed (\textbf{speed error}). 
We sample 14 scenarios, one of each scenario type in nuPlan, and report average planning costs across all scenarios.

\paragraph{Baselines}
We consider the following baselines: 
\begin{itemize}
    \item \textit{CEM} \cite{rubinstein1997optimization}: a widely used ES method which parameterizes the search distribution $q$ as a Gaussian.
    CEM iterates between sampling from $q$ and fitting $q$ to the best samples.
    \item \textit{MPPI} \cite{williams2015model}: similar to CEM but rather than keep a fixed number of elites, MPPI samples proportional to rewards.
    \item \textit{Reward-gradient guidance} \cite{janner2022diffuser}: we directly optimize the ground-truth planning objective with gradient descent during the denoising process.
\end{itemize}

\begin{table}
\begin{center}
\begin{tabular}{l|c|c}
\hline
Method & Lane Error & Speed Error\\
\hline
CEM & 2.34 & 2.05 \\
MPPI & 3.29 & 2.74 \\
Reward-gradient guidance  & 1.22 & 0.96 \\
\model{} (ours) &\textbf{ 0.61} & \textbf{0.79} \\
\hline
\end{tabular}
\end{center}
\caption{\textbf{Lane following in nuPlan.}
}
\label{tab:lanefollowing}
\end{table}

We show quantitative results  in Table  \ref{tab:lanefollowing}. 
We draw the following conclusions: 
\noindent \textbf{1.\model{} outperforms the differentiable reward-gradient guidance baseline \textit{even though the objective is differentiable}}. 
We hypothesize that this is because although the ground truth reward function is available, it may not provide suitable guidance for intermediate noisy trajectories.  
This highlights a key advantage of our method over prior work, which is that we can optimize novel objectives without needing to train a reward regressor on noisy samples.
\noindent \textbf{2. Both diffusion-based methods significantly outperform sampling-based planners that do not leverage diffusion}.
This is consistent with our hypothesis that diffusion guidance can optimize trajectories much more efficiently than conventional ES methods. 
Videos of our method driving in all experimental settings can be found in our project page \url{diffusion-es.github.io}.

\subsection{Runtime analysis}
\label{sec:runtime}
\begin{table}
\begin{center}
\begin{tabular}{l|c}
\hline
Method & Wallclock time (s) \\
\hline
Diffusion  & 1.11 $\pm$ 0.02\\
\model{} & 5.85 $\pm$ 0.11\\
\model{} (\textit{optimized}) & 0.50 $\pm$ 0.01\\
\hline
\end{tabular}
\end{center}
\caption{\textbf{Diffusion-ES runtime comparison. }
}
\label{tab:runtime}
\end{table}
\model{} can be used in real-time with some minor optimizations.
By using fewer diffusion steps $T=10$, smaller population size $M=32$ and less iterations $K=2$, our method can be run at the same frequency as the simulator (2 Hz) at a small cost to performance (nuPlan driving score drops from 92 to 91).
We report the average wallclock time for inference at every timestep over 100 trials in Table \ref{tab:runtime}.


\subsection{Discussion - Limitations - Future work}
As seen in Section \ref{sec:runtime}, our approach does introduce computational overhead. 
We believe that these issues can be mitigated by incorporating recent advances in diffusion modeling such as faster samplers.
Our reward function assumes other agents will travel at constant velocity, which could clearly be improved.
This also assumes that other agents cannot react to the ego-vehicle, which has been shown to be a major limitation for planners in self-driving \cite{rhinehart2021contingencies}.
However, our instruction following experiments suggest that even if we cannot ever forecast perfectly, we can use language-shaped rewards to solve the hardest driving scenarios.
We aim to explore memory-prompted analogical reward shaping for handling long-tail scenarios without a human teacher in our future work.

%% file: 6_conclusion.tex
\section{Conclusion}
We presented \model{}, a method for black-box reward guided diffusion sampling.
We showed that  \model{} can effectively optimize reward functions in nuPlan for driving and instruction following, and outperforms engineered sampling-based planners, reactive deterministic or diffusion policies, as well as differentiable reward-gradient guidance.
We showed how our method can be used to follow language instructions without any language-action trajectory data, simply using LLM prompting to generate shaped reward maps for test-time optimization.  
Our future work will explore retrieving the right reward shaping to optimize to handle long-tailed driving scenarios in the absence of human teachers. 
Our experiments show the trade-off between inference speed and OOD generalization during scene conditioning in diffusion  policies.
Our future work will explore ways to amortize the result of such searches to fast reactive policies, and to balance the two extremes so that a variable amount of compute can be spent depending on the scenario. 

%% file: 7_acknowledgements.tex
\section*{Acknowledgements}

This material is based upon work supported by  an NSF CAREER award, AFOSR FA95502310257 award, an ONR award N000142312415, and ARL award No 48987.3.1130233. Any opinions, findings and conclusions or recommendations expressed in this material are those of the authors and do not necessarily reflect the views of the United States Army, the National Science Foundation, or the United States Air Force.

%% file: 8_appendix.tex
\newpage
\clearpage
\section{Appendix}

\subsection{Model training details}
We train our diffusion models on 1\% of the released nuPlan training data, which is subsampled from the original 20Hz to 0.5Hz.
Our diffusion models are DDIMs \cite{song2020denoising} trained with $T=100$ diffusion steps. We use the scaled linear beta schedule and predict $\epsilon$.
Our models are implemented in PyTorch and we use the HuggingFace Diffusers library to implement our diffusion models.

Our base diffusion architecture is as follows.
Each trajectory waypoint is linearly projected to a latent feature with hidden size 256.
The noise level is encoded with sinusoidal positional embeddings followed by a 2-layer MLP.
Noise features are fused with the trajectory tokens by concatenating the noise feature to all trajectory features along the feature dimension and projecting back to the hidden size of 256.
We also apply rotary positional embeddings \cite{su2021roformer} to the trajectory tokens as temporal embeddings.
We then pass all the trajectory tokens through 8 transformer encoder layers, and each trajectory token is decoded to a corresponding waypoint.
The final trajectory consists of the stacked waypoint predictions.
Our conditional diffusion policy baseline uses a similar architecture, except we featurize the scene using the backbone from the nuPlan re-implementation of Urban Driver \cite{scheel2022urban} and pass those tokens into the self-attention layers.

We train our models with batch size 256 and use the AdamW optimizer with learning rate 1e-4, weight decay 5e-4, and $(\beta_1, \beta_2) = (0.9, 0.999)$.

Our trajectories consist of 16 2D pose waypoints each with 3 features ($x,y,\theta$).
We preprocess these trajectory features by applying Verlet wrapping as described in MotionLM \cite{seff2023motionlm}, which we found to improve performance by encouraging smooth trajectories.

\subsection{Language instruction following tasks}
Here, we describe in detail each of the controllability tasks.
For each, we list the task goal as well as the specific language instruction used.
\begin{enumerate}
    \item \textbf{Lane change}: the ego-vehicle must execute a lane change. The language instruction is \textit{"Change lanes to the left"}. The episode is considered a success if the ego-vehicle reaches the left lane.
    \item \textbf{Unprotected left turn}: the ego-vehicle must perform an unprotected left turn. The language instruction is \textit{"If car 18 is within 20 meters yield to it. Otherwise it will slow for you"}, where car 18 is the incoming car. The episode is considered a success if the ego-vehicle either completes the turn before the incoming car, or the incoming car passes the ego freely indicating a successful yield. Due to the randomized agent behaviors, it is not always possible to safely execute the turn in this task.
    \item \textbf{Unprotected right turn}: the ego-vehicle must perform an unprotected right turn. The language instruction is \textit{"Change to lane 33."}, where lane 33 is the target lane. The episode is considered a success if the ego-vehicle completes the right turn.
    \item \textbf{Overtaking}: the ego-vehicle must overtake the target car. The language instruction is "Car 21 will slow for you. Change to the right lane. Once ahead of car 4 change to the left lane.", where car 21 is the incoming car in the right lane and car 4 is the target car. The episode is considered a success if the ego-vehicle is ahead of the target car while in the same lane.
    \item \textbf{Extended overtaking}: the ego-vehicle must overtake the target car across several lanes of dense traffic. The language instruction is \textit{"Change two lanes to the left. Then if you are ever ahead of car 3 change lanes to the right"}. The episode is considered a success if the ego-vehicle is ahead of the target car while in the same lane.
    \item \textbf{Yielding}: the ego-vehicle must allow a car approaching quickly from behind to pass by changing lanes. The language instruction is \textit{"Slow down and change lanes to the left. Then once car 8 is ahead of you change lanes to the right."}, where car 8 is the incoming car. The episode is considered a success if the ego-vehicle is behind the target car.
    \item \textbf{Cut in}: the ego-vehicle must cut in to a column of cars. The language instruction is \textit{"Slow down. Car 2 will slow for you. Change two lanes to the left"}, where car 2 is the car we are cutting in front of. The episode is considered a success if the ego-vehicle is in front of the target car and in the same lane.
    \item \textbf{Lane weaving}: the ego-vehicle must reach a specific gap between two cars across several lanes of dense traffic. The language instruction is \textit{"Once ahead of car 12 change lanes to the left. Then slow down a lot and change lanes to the left. Once car 9 is ahead of you by a few meters change lanes to the left"}. The episode is considered a success if the ego-vehicle successfully reaches the target gap.
\end{enumerate}

\subsection{Language instruction following prompts}
We collect 24 instruction-program pairs in total, and they are shown in full below.
To organize and manage our prompts, we use DSPy \cite{khattab2023dspy}.

\label{prompt1}
\onecolumn\begin{lstlisting}[caption={Full prompt with all instruction-program pairs},captionpos=t,label={prompt1}] 
Use the provided language instruction to write code for guiding a lower-level driving policy.

---

Follow the following format.

Instruction: ${instruction}
Code: ${code}

---

Instruction: Yield to car 4. If car 2 is ahead of car 4 stop yielding.
Code:
done = self.yield_to_vehicle(self.get_vehicle(4))
def vehicle_ahead_of_other_vehicle():
    vehicle = self.get_vehicle(2)
    other_vehicle = self.get_vehicle(4)
    return vehicle.is_ahead_of(other_vehicle)
while not done() and not vehicle_ahead_of_other_vehicle():
    yield
self.stop_yielding()

---

Instruction: If car 1 is not stopped yield to it.
Code:
vehicle = self.get_vehicle(1)
if not vehicle.is_stopped():
    done = self.yield_to_vehicle(vehicle)
    while not done():
        yield
    self.stop_yielding()

---

Instruction: Change to the left lane.
Code:
done = self.follow_lane(self.left_lane)
while not done():
    yield

---

Instruction: If the speed of car 2 is lower than the speed limit change to the right lane.
Code:
vehicle = self.get_vehicle(2)
if vehicle.speed < self.current_lane.speed_limit:
    done = self.follow_lane(self.right_lane)
    while not done():
        yield

---

Instruction: Car 20 will slow down.
Code:
vehicle = self.get_vehicle(20)
self.adjust_constant_velocity_prediction(vehicle, 0.5)

---

Instruction: If car 17 is within 20 meters yield to it.
Code:
vehicle = self.get_vehicle(17)
if vehicle.distance_to(self.ego_vehicle) < 20.0:
    done = self.yield_to_vehicle(vehicle)
    while not done():
        yield
    self.stop_yielding()

---

Instruction: Resume normal driving speed.
Code: self.unset_ego_speed_limit()

---

Instruction: While you are ahead of car 3 stay in the current lane. Otherwise change to their lane.
Code:
def ahead_of_vehicle():
    vehicle = self.get_vehicle(3)
    return self.ego_vehicle.is_ahead_of(vehicle)
while ahead_of_vehicle():
    self.follow_lane(self.current_lane)
    yield
vehicle = self.get_vehicle(3)
their_lane = vehicle.get_closest_lane(self.lane_graph)
done = self.follow_lane(their_lane)
while not done():
    yield

---

Instruction: Follow lane 12.
Code:
lane = self.get_lane(12)
done = self.follow_lane(lane)
while not done():
    yield

---

Instruction: Once the speed of car 2 is lower than the speed limit of your current lane change to the right lane.
Code:
speed_limit = self.current_lane.speed_limit
def speed_exceeds_limit():
    vehicle = self.get_vehicle(2)
    return vehicle.speed < speed_limit
while speed_exceeds_limit():
    yield
done = self.follow_lane(self.right_lane)
while not done():
    yield

---

Instruction: Change to lane 1.
Code:
lane = self.get_lane(1)
done = self.follow_lane(lane)
while not done():
    yield

---

Instruction: Yield to car 3.
Code:
vehicle = self.get_vehicle(3)
done = self.yield_to_vehicle(vehicle)
while not done():
    yield
self.stop_yielding()

---

Instruction: If car 2 is ever ahead of you by 10 meters it will slow for you.
Code:
def vehicle_ahead_of_us():
    vehicle = self.get_vehicle(2)
    return vehicle.is_ahead_of(self.ego_vehicle, 10.0)
while not vehicle_ahead_of_us():
    yield
self.set_velocity_ratio(self.get_vehicle(2), 0.5)

---

Instruction: Change lanes to the right. Then change lanes to the left.
Code:
done = self.follow_lane(self.right_lane)
while not done():
    yield
done = self.follow_lane(self.left_lane)
while not done():
    yield

---

Instruction: If the speed of car 8 is greater than 5.0 yield to it.
Code:
vehicle = self.get_vehicle(8)
if vehicle.speed > 5.0:
    done = self.yield_to_vehicle(vehicle)
    while not done():
        yield
    self.stop_yielding()

---

Instruction: Change lanes to the right. Once car 1 is ahead of you yield to it.
Code:
done = self.follow_lane(self.right_lane)
while not done():
    yield
def vehicle_ahead_of_us():
    vehicle = self.get_vehicle(1)
    return vehicle.is_ahead_of(self.ego_vehicle)
while not vehicle_ahead_of_us():
    yield
done = self.yield_to_vehicle(self.get_vehicle(1))
while not done():
    yield
self.stop_yielding()

---

Instruction: Slow down and change lanes to the right.
Code:
current_speed = self.ego_vehicle.speed
self.set_ego_speed_limit(current_speed * 0.5)
done = self.follow_lane(self.right_lane)
while not done():
    yield

---

Instruction: Slow down a lot and change lanes to the right. Then once car 2 is ahead of you, resume normal driving speed.
Code:
current_speed = self.ego_vehicle.speed
self.set_ego_speed_limit(current_speed * 0.2)
done = self.follow_lane(self.right_lane)
while not done():
    yield
def vehicle_ahead_of_us():
    vehicle = self.get_vehicle(2)
    return vehicle.is_ahead_of(self.ego_vehicle)
while not vehicle_ahead_of_us():
    yield
current_speed = self.ego_vehicle.speed
self.unset_ego_speed_limit()

---

Instruction: Stay in the current lane. If your speed exceeds 2.0 stop following the current lane.
Code:
done = self.follow_lane(self.current_lane)
def speed_under_threshold():
    speed = self.ego_vehicle.speed
    return speed < 2.0
while not done() and speed_under_threshold():
    yield
self.stop_following()

---

Instruction: While car 3 is ahead of you stay in the current lane. Otherwise change to their lane.
Code:
def vehicle_ahead_of_us():
    vehicle = self.get_vehicle(3)
    return vehicle.is_ahead_of(self.ego_vehicle)
while vehicle_ahead_of_us():
    self.follow_lane(self.current_lane)
    yield
vehicle = self.get_vehicle(3)
their_lane = vehicle.get_closest_lane(self.lane_graph)
done = self.follow_lane(their_lane)
while not done():
    yield

---

Instruction: If there is a car in the left lane going faster than 5.0 m/s stay in the current lane.
Code:
vehicles_in_left_lane = self.left_lane.get_vehicles()
if any([vehicle.speed > 5.0 for vehicle in vehicles_in_left_lane]):
    done = self.follow_lane(self.current_lane)
    while not done():
        yield

---

Instruction: If car 2 is ahead of you by 10 meters it will slow for you.
Code:
vehicle = self.get_vehicle(2)
if vehicle.is_ahead_of(self.ego_vehicle, 10.0):
    self.adjust_constant_velocity_prediction(vehicle, 0.5)

---

Instruction: Stay in the current lane.
Code:
done = self.follow_lane(self.current_lane)
while not done():
    yield

---

Instruction: If car 20 is ever stopped yield to it.
Code:
def vehicle_is_ever_stopped():
    vehicle = self.get_vehicle(20)
    return vehicle.is_stopped()
while not vehicle_is_ever_stopped():
    yield
vehicle = self.get_vehicle(20)
done = self.yield_to_vehicle(vehicle)
while not done():
    yield
self.stop_yielding()

---

Instruction: [insert command here]
Code: 
\end{lstlisting}

%% file: 00_main.bbl
\begin{thebibliography}{77}
\providecommand{\natexlab}[1]{#1}
\providecommand{\url}[1]{\texttt{#1}}
\expandafter\ifx\csname urlstyle\endcsname\relax
  \providecommand{\doi}[1]{doi: #1}\else
  \providecommand{\doi}{doi: \begingroup \urlstyle{rm}\Url}\fi

\bibitem[Ajay et~al.(2022)Ajay, Du, Gupta, Tenenbaum, Jaakkola, and Agrawal]{Ajay2022IsCG}
Anurag Ajay, Yilun Du, Abhi Gupta, Joshua~B. Tenenbaum, T. Jaakkola, and Pulkit Agrawal.
\newblock Is conditional generative modeling all you need for decision-making?
\newblock \emph{ArXiv}, abs/2211.15657, 2022.

\bibitem[Bansal et~al.(2018)Bansal, Krizhevsky, and Ogale]{chauffernet}
Mayank Bansal, Alex Krizhevsky, and Abhijit~S. Ogale.
\newblock Chauffeurnet: Learning to drive by imitating the best and synthesizing the worst.
\newblock \emph{CoRR}, abs/1812.03079, 2018.

\bibitem[Bojarski et~al.(2016)Bojarski, Del~Testa, Dworakowski, Firner, Flepp, Goyal, Jackel, Monfort, Muller, Zhang, Zhang, Zhao, and Zieba]{endtoendNVIDIA}
Mariusz Bojarski, Davide Del~Testa, Daniel Dworakowski, Bernhard Firner, Beat Flepp, Prasoon Goyal, Lawrence~D. Jackel, Mathew Monfort, Urs Muller, Jiakai Zhang, Xin Zhang, Jake Zhao, and Karol Zieba.
\newblock End to end learning for self-driving cars, 2016.

\bibitem[Brown et~al.(2020)Brown, Mann, Ryder, Subbiah, Kaplan, Dhariwal, Neelakantan, Shyam, Sastry, Askell, Agarwal, Herbert-Voss, Krueger, Henighan, Child, Ramesh, Ziegler, Wu, Winter, Hesse, Chen, Sigler, Litwin, Gray, Chess, Clark, Berner, McCandlish, Radford, Sutskever, and Amodei]{brown2020language}
Tom~B. Brown, Benjamin Mann, Nick Ryder, Melanie Subbiah, Jared Kaplan, Prafulla Dhariwal, Arvind Neelakantan, Pranav Shyam, Girish Sastry, Amanda Askell, Sandhini Agarwal, Ariel Herbert-Voss, Gretchen Krueger, Tom Henighan, Rewon Child, Aditya Ramesh, Daniel~M. Ziegler, Jeffrey Wu, Clemens Winter, Christopher Hesse, Mark Chen, Eric Sigler, Mateusz Litwin, Scott Gray, Benjamin Chess, Jack Clark, Christopher Berner, Sam McCandlish, Alec Radford, Ilya Sutskever, and Dario Amodei.
\newblock Language models are few-shot learners, 2020.

\bibitem[Caesar et~al.(2022)Caesar, Kabzan, Tan, Fong, Wolff, Lang, Fletcher, Beijbom, and Omari]{caesar2022nuplan}
Holger Caesar, Juraj Kabzan, Kok~Seang Tan, Whye~Kit Fong, Eric Wolff, Alex Lang, Luke Fletcher, Oscar Beijbom, and Sammy Omari.
\newblock Nuplan: A closed-loop ml-based planning benchmark for autonomous vehicles, 2022.

\bibitem[Carvalho et~al.(2023)Carvalho, Le, Baierl, Koert, and Peters]{Carvalho2023MotionPD}
Jo{\~a}o Carvalho, An~T. Le, Mark Baierl, Dorothea Koert, and Jan Peters.
\newblock Motion planning diffusion: Learning and planning of robot motions with diffusion models.
\newblock \emph{ArXiv}, abs/2308.01557, 2023.

\bibitem[Chen and Krähenbühl(2022)]{LAV}
Dian Chen and Philipp Krähenbühl.
\newblock Learning from all vehicles, 2022.

\bibitem[Chen et~al.(2019)Chen, Zhou, Koltun, and Kr{\"a}henb{\"u}hl]{LBC}
Dian Chen, Brady Zhou, Vladlen Koltun, and Philipp Kr{\"a}henb{\"u}hl.
\newblock Learning by cheating.
\newblock \emph{ArXiv}, abs/1912.12294, 2019.

\bibitem[Chi et~al.(2023)Chi, Feng, Du, Xu, Cousineau, Burchfiel, and Song]{chi2023diffusion}
Cheng Chi, Siyuan Feng, Yilun Du, Zhenjia Xu, Eric Cousineau, Benjamin Burchfiel, and Shuran Song.
\newblock Diffusion policy: Visuomotor policy learning via action diffusion, 2023.

\bibitem[Codevilla et~al.(2017)Codevilla, M{\"{u}}ller, Dosovitskiy, L{\'{o}}pez, and Koltun]{DBLP:journals/corr/abs-1710-02410}
Felipe Codevilla, Matthias M{\"{u}}ller, Alexey Dosovitskiy, Antonio~M. L{\'{o}}pez, and Vladlen Koltun.
\newblock End-to-end driving via conditional imitation learning.
\newblock \emph{CoRR}, abs/1710.02410, 2017.

\bibitem[Cui et~al.(2021)Cui, Sadat, Casas, Liao, and Urtasun]{lookout}
Alexander Cui, Abbas Sadat, Sergio Casas, Renjie Liao, and Raquel Urtasun.
\newblock Lookout: Diverse multi-future prediction and planning for self-driving.
\newblock \emph{CoRR}, abs/2101.06547, 2021.

\bibitem[Dauner et~al.(2023)Dauner, Hallgarten, Geiger, and Chitta]{dauner2023parting}
Daniel Dauner, Marcel Hallgarten, Andreas Geiger, and Kashyap Chitta.
\newblock Parting with misconceptions about learning-based vehicle motion planning, 2023.

\bibitem[Dhariwal and Nichol(2021)]{Dhariwal2021DiffusionMB}
Prafulla Dhariwal and Alex Nichol.
\newblock Diffusion models beat gans on image synthesis.
\newblock \emph{ArXiv}, abs/2105.05233, 2021.

\bibitem[Du et~al.(2023{\natexlab{a}})Du, Durkan, Strudel, Tenenbaum, Dieleman, Fergus, Sohl-Dickstein, Doucet, and Grathwohl]{du2023reduce}
Yilun Du, Conor Durkan, Robin Strudel, Joshua~B. Tenenbaum, Sander Dieleman, Rob Fergus, Jascha Sohl-Dickstein, Arnaud Doucet, and Will Grathwohl.
\newblock Reduce, reuse, recycle: Compositional generation with energy-based diffusion models and mcmc, 2023{\natexlab{a}}.

\bibitem[Du et~al.(2023{\natexlab{b}})Du, Yang, Dai, Dai, Nachum, Tenenbaum, Schuurmans, and Abbeel]{Du2023LearningUP}
Yilun Du, Mengjiao Yang, Bo Dai, Hanjun Dai, Ofir Nachum, Joshua~B. Tenenbaum, Dale Schuurmans, and P. Abbeel.
\newblock Learning universal policies via text-guided video generation.
\newblock \emph{ArXiv}, abs/2302.00111, 2023{\natexlab{b}}.

\bibitem[Fan et~al.(2018)Fan, Zhu, Liu, Zhang, Zhuang, Li, Zhu, Hu, Li, and Kong]{fan2018baidu}
Haoyang Fan, Fan Zhu, Changchun Liu, Liangliang Zhang, Li Zhuang, Dong Li, Weicheng Zhu, Jiangtao Hu, Hongye Li, and Qi Kong.
\newblock Baidu apollo em motion planner.
\newblock \emph{arXiv preprint arXiv:1807.08048}, 2018.

\bibitem[Fragkiadaki et~al.(2017)Fragkiadaki, Huang, Alemi, Vijayanarasimhan, Ricco, and Sukthankar]{fragkiadaki2017motion}
Katerina Fragkiadaki, Jonathan Huang, Alex Alemi, Sudheendra Vijayanarasimhan, Susanna Ricco, and Rahul Sukthankar.
\newblock Motion prediction under multimodality with conditional stochastic networks, 2017.

\bibitem[Gkanatsios et~al.(2023)Gkanatsios, Jain, Xian, Zhang, Atkeson, and Fragkiadaki]{gkanatsios2023energybased}
Nikolaos Gkanatsios, Ayush Jain, Zhou Xian, Yunchu Zhang, Christopher Atkeson, and Katerina Fragkiadaki.
\newblock {Energy-based Models are Zero-Shot Planners for Compositional Scene Rearrangement}.
\newblock In \emph{Robotics: Science and Systems}, 2023.

\bibitem[Gupta and Kembhavi(2023)]{Gupta_2023_CVPR}
Tanmay Gupta and Aniruddha Kembhavi.
\newblock Visual programming: Compositional visual reasoning without training.
\newblock In \emph{Proceedings of the IEEE/CVF Conference on Computer Vision and Pattern Recognition (CVPR)}, pages 14953--14962, 2023.

\bibitem[Helbing and Tilch(1998)]{helbing1998generalized}
Dirk Helbing and Benno Tilch.
\newblock Generalized force model of traffic dynamics.
\newblock \emph{Physical review E}, 58\penalty0 (1):\penalty0 133, 1998.

\bibitem[Ho et~al.(2020)Ho, Jain, and Abbeel]{ho2020denoising}
Jonathan Ho, Ajay Jain, and Pieter Abbeel.
\newblock Denoising diffusion probabilistic models.
\newblock \emph{Advances in Neural Information Processing Systems}, 33:\penalty0 6840--6851, 2020.

\bibitem[Hu et~al.(2023)Hu, Russell, Yeo, Murez, Fedoseev, Kendall, Shotton, and Corrado]{hu2023gaia}
Anthony Hu, Lloyd Russell, Hudson Yeo, Zak Murez, George Fedoseev, Alex Kendall, Jamie Shotton, and Gianluca Corrado.
\newblock Gaia-1: A generative world model for autonomous driving.
\newblock \emph{arXiv preprint arXiv:2309.17080}, 2023.

\bibitem[Huang et~al.(2022{\natexlab{a}})Huang, Abbeel, Pathak, and Mordatch]{huang2022language}
Wenlong Huang, Pieter Abbeel, Deepak Pathak, and Igor Mordatch.
\newblock Language models as zero-shot planners: Extracting actionable knowledge for embodied agents.
\newblock \emph{arXiv preprint arXiv:2201.07207}, 2022{\natexlab{a}}.

\bibitem[Huang et~al.(2022{\natexlab{b}})Huang, Xia, Xiao, Chan, Liang, Florence, Zeng, Tompson, Mordatch, Chebotar, Sermanet, Brown, Jackson, Luu, Levine, Hausman, and Ichter]{https://doi.org/10.48550/arxiv.2207.05608}
Wenlong Huang, Fei Xia, Ted Xiao, Harris Chan, Jacky Liang, Pete Florence, Andy Zeng, Jonathan Tompson, Igor Mordatch, Yevgen Chebotar, Pierre Sermanet, Noah Brown, Tomas Jackson, Linda Luu, Sergey Levine, Karol Hausman, and Brian Ichter.
\newblock Inner monologue: Embodied reasoning through planning with language models, 2022{\natexlab{b}}.

\bibitem[Huang et~al.(2023)Huang, Wang, Zhang, Li, Wu, and Fei-Fei]{huang2023voxposer}
Wenlong Huang, Chen Wang, Ruohan Zhang, Yunzhu Li, Jiajun Wu, and Li Fei-Fei.
\newblock Voxposer: Composable 3d value maps for robotic manipulation with language models.
\newblock \emph{arXiv preprint arXiv:2307.05973}, 2023.

\bibitem[Janner et~al.(2022)Janner, Du, Tenenbaum, and Levine]{janner2022diffuser}
Michael Janner, Yilun Du, Joshua Tenenbaum, and Sergey Levine.
\newblock Planning with diffusion for flexible behavior synthesis.
\newblock In \emph{International Conference on Machine Learning}, 2022.

\bibitem[Jiang et~al.(2023{\natexlab{a}})Jiang, Cornman, Park, Sapp, Zhou, and Anguelov]{Jiang_2023_CVPR}
Chiyu Jiang, Andre Cornman, Cheolho Park, Benjamin Sapp, Yin Zhou, and Dragomir Anguelov.
\newblock Motiondiffuser: Controllable multi-agent motion prediction using diffusion.
\newblock In \emph{Proceedings of the IEEE/CVF Conference on Computer Vision and Pattern Recognition (CVPR)}, pages 9644--9653, 2023{\natexlab{a}}.

\bibitem[Jiang et~al.(2023{\natexlab{b}})Jiang, Cornman, Park, Sapp, Zhou, and Anguelov]{MotionDiffuser}
Chiyu~“Max” Jiang, Andre Cornman, Cheolho Park, Benjamin Sapp, Yin Zhou, and Dragomir Anguelov.
\newblock Motiondiffuser: Controllable multi-agent motion prediction using diffusion, 2023{\natexlab{b}}.

\bibitem[Katara et~al.(2023)Katara, Xian, and Fragkiadaki]{gen2sim}
Pushkal Katara, Zhou Xian, and Katerina Fragkiadaki.
\newblock Gen2sim: Scaling up robot learning in simulation with generative models, 2023.

\bibitem[Khattab et~al.(2023)Khattab, Singhvi, Maheshwari, Zhang, Santhanam, Vardhamanan, Haq, Sharma, Joshi, Moazam, Miller, Zaharia, and Potts]{khattab2023dspy}
Omar Khattab, Arnav Singhvi, Paridhi Maheshwari, Zhiyuan Zhang, Keshav Santhanam, Sri Vardhamanan, Saiful Haq, Ashutosh Sharma, Thomas~T. Joshi, Hanna Moazam, Heather Miller, Matei Zaharia, and Christopher Potts.
\newblock Dspy: Compiling declarative language model calls into self-improving pipelines.
\newblock \emph{arXiv preprint arXiv:2310.03714}, 2023.

\bibitem[Kim et~al.(2023)Kim, Jeon, Choi, and Kum]{Kim_2023}
Sanmin Kim, Hyeongseok Jeon, Jun~Won Choi, and Dongsuk Kum.
\newblock Diverse multiple trajectory prediction using a two-stage prediction network trained with lane loss.
\newblock \emph{{IEEE} Robotics and Automation Letters}, 8\penalty0 (4):\penalty0 2038--2045, 2023.

\bibitem[Kuefler et~al.(2017)Kuefler, Morton, Wheeler, and Kochenderfer]{gaildriving}
Alex Kuefler, Jeremy Morton, Tim~Allan Wheeler, and Mykel~John Kochenderfer.
\newblock Imitating driver behavior with generative adversarial networks.
\newblock \emph{CoRR}, abs/1701.06699, 2017.

\bibitem[Liang et~al.(2022)Liang, Huang, Xia, Xu, Hausman, Ichter, Florence, and Zeng]{liang2022code}
Jacky Liang, Wenlong Huang, Fei Xia, Peng Xu, Karol Hausman, Brian Ichter, Pete Florence, and Andy Zeng.
\newblock Code as policies: Language model programs for embodied control.
\newblock \emph{arXiv preprint arXiv:2209.07753}, 2022.

\bibitem[Liang et~al.(2023)Liang, Mu, Ding, Ni, Tomizuka, and Luo]{Liang2023AdaptDiffuserDM}
Zhixuan Liang, Yao Mu, Mingyu Ding, Fei Ni, Masayoshi Tomizuka, and Ping Luo.
\newblock Adaptdiffuser: Diffusion models as adaptive self-evolving planners.
\newblock In \emph{ICML}, 2023.

\bibitem[Liu et~al.(2021)Liu, Yuan, Fu, Jiang, Hayashi, and Neubig]{DBLP:journals/corr/abs-2107-13586}
Pengfei Liu, Weizhe Yuan, Jinlan Fu, Zhengbao Jiang, Hiroaki Hayashi, and Graham Neubig.
\newblock Pre-train, prompt, and predict: {A} systematic survey of prompting methods in natural language processing.
\newblock \emph{CoRR}, abs/2107.13586, 2021.

\bibitem[Lu et~al.(2023)Lu, Chen, Chen, Su, Li, and Zhu]{lu2023contrastive}
Cheng Lu, Huayu Chen, Jianfei Chen, Hang Su, Chongxuan Li, and Jun Zhu.
\newblock Contrastive energy prediction for exact energy-guided diffusion sampling in offline reinforcement learning.
\newblock \emph{arXiv preprint arXiv:2304.12824}, 2023.

\bibitem[Mao et~al.(2023)Mao, Qian, Zhao, and Wang]{mao2023gpt}
Jiageng Mao, Yuxi Qian, Hang Zhao, and Yue Wang.
\newblock Gpt-driver: Learning to drive with gpt.
\newblock \emph{arXiv preprint arXiv:2310.01415}, 2023.

\bibitem[Montremerlo et~al.(2008)Montremerlo, Beeker, Bhat, and Dahlkamp]{montremerlo2008stanford}
M Montremerlo, J Beeker, S Bhat, and H Dahlkamp.
\newblock The stanford entry in the urban challenge.
\newblock \emph{Journal of Field Robotics}, 7\penalty0 (9):\penalty0 468--492, 2008.

\bibitem[Nayakanti et~al.(2022)Nayakanti, Al-Rfou, Zhou, Goel, Refaat, and Sapp]{wayformer}
Nigamaa Nayakanti, Rami Al-Rfou, Aurick Zhou, Kratarth Goel, Khaled~S. Refaat, and Benjamin Sapp.
\newblock Wayformer: Motion forecasting via simple and efficient attention networks, 2022.

\bibitem[Nichol and Dhariwal(2021)]{nichol2021improved}
Alexander~Quinn Nichol and Prafulla Dhariwal.
\newblock Improved denoising diffusion probabilistic models.
\newblock In \emph{International Conference on Machine Learning}, pages 8162--8171. PMLR, 2021.

\bibitem[Pearce et~al.(2023)Pearce, Rashid, Kanervisto, Bignell, Sun, Georgescu, Macua, Tan, Momennejad, Hofmann, and Devlin]{pearce2023imitating}
Tim Pearce, Tabish Rashid, Anssi Kanervisto, Dave Bignell, Mingfei Sun, Raluca Georgescu, Sergio~Valcarcel Macua, Shan~Zheng Tan, Ida Momennejad, Katja Hofmann, and Sam Devlin.
\newblock Imitating human behaviour with diffusion models, 2023.

\bibitem[Polack et~al.(2017)Polack, Altch{\'e}, d'Andr{\'e}a Novel, and de~La~Fortelle]{polack2017kinematic}
Philip Polack, Florent Altch{\'e}, Brigitte d'Andr{\'e}a Novel, and Arnaud de La~Fortelle.
\newblock The kinematic bicycle model: A consistent model for planning feasible trajectories for autonomous vehicles?
\newblock In \emph{2017 IEEE intelligent vehicles symposium (IV)}, pages 812--818. IEEE, 2017.

\bibitem[Pomerleau(1989)]{alvinn}
Dean~A. Pomerleau.
\newblock {ALVINN:} an autonomous land vehicle in a neural network.
\newblock In \emph{Advances in Neural Information Processing Systems 1}, pages 305--313. San Francisco, CA: Morgan Kaufmann, 1989.

\bibitem[Prakash et~al.(2021)Prakash, Chitta, and Geiger]{transfuser}
Aditya Prakash, Kashyap Chitta, and Andreas Geiger.
\newblock Multi-modal fusion transformer for end-to-end autonomous driving.
\newblock \emph{CoRR}, abs/2104.09224, 2021.

\bibitem[Ramesh et~al.(2022)Ramesh, Dhariwal, Nichol, Chu, and Chen]{Ramesh2022HierarchicalTI}
Aditya Ramesh, Prafulla Dhariwal, Alex Nichol, Casey Chu, and Mark Chen.
\newblock Hierarchical text-conditional image generation with clip latents.
\newblock \emph{ArXiv}, abs/2204.06125, 2022.

\bibitem[Renz et~al.(2022)Renz, Chitta, Mercea, Koepke, Akata, and Geiger]{plant}
Katrin Renz, Kashyap Chitta, Otniel-Bogdan Mercea, A.~Sophia Koepke, Zeynep Akata, and Andreas Geiger.
\newblock Plant: Explainable planning transformers via object-level representations.
\newblock In \emph{6th Annual Conference on Robot Learning}, 2022.

\bibitem[Reuss et~al.(2023)Reuss, Li, Jia, and Lioutikov]{reuss2023goal}
Moritz Reuss, Maximilian Li, Xiaogang Jia, and Rudolf Lioutikov.
\newblock Goal-conditioned imitation learning using score-based diffusion policies.
\newblock \emph{arXiv preprint arXiv:2304.02532}, 2023.

\bibitem[Rhinehart et~al.(2021)Rhinehart, He, Packer, Wright, McAllister, Gonzalez, and Levine]{rhinehart2021contingencies}
Nicholas Rhinehart, Jeff He, Charles Packer, Matthew~A Wright, Rowan McAllister, Joseph~E Gonzalez, and Sergey Levine.
\newblock Contingencies from observations: Tractable contingency planning with learned behavior models.
\newblock In \emph{2021 IEEE International Conference on Robotics and Automation (ICRA)}, pages 13663--13669. IEEE, 2021.

\bibitem[Rombach et~al.(2022)Rombach, Blattmann, Lorenz, Esser, and Ommer]{Rombach2021HighResolutionIS}
Robin Rombach, A. Blattmann, Dominik Lorenz, Patrick Esser, and Bj{\"o}rn Ommer.
\newblock High-resolution image synthesis with latent diffusion models.
\newblock \emph{2022 IEEE/CVF Conference on Computer Vision and Pattern Recognition (CVPR)}, pages 10674--10685, 2022.

\bibitem[Rubinstein(1997)]{rubinstein1997optimization}
Reuven~Y Rubinstein.
\newblock Optimization of computer simulation models with rare events.
\newblock \emph{European Journal of Operational Research}, 99\penalty0 (1):\penalty0 89--112, 1997.

\bibitem[Saharia et~al.(2022)Saharia, Chan, Saxena, Li, Whang, Denton, Ghasemipour, Ayan, Mahdavi, Lopes, Salimans, Ho, Fleet, and Norouzi]{Saharia2022PhotorealisticTD}
Chitwan Saharia, William Chan, Saurabh Saxena, Lala Li, Jay Whang, Emily~L. Denton, Seyed Kamyar~Seyed Ghasemipour, Burcu~Karagol Ayan, Seyedeh~Sara Mahdavi, Raphael~Gontijo Lopes, Tim Salimans, Jonathan Ho, David~J. Fleet, and Mohammad Norouzi.
\newblock Photorealistic text-to-image diffusion models with deep language understanding.
\newblock \emph{ArXiv}, abs/2205.11487, 2022.

\bibitem[Scheel et~al.(2021)Scheel, Bergamini, Wołczyk, Osi'nski, and Ondruska]{Scheel2021UrbanDL}
Oliver Scheel, Luca Bergamini, Maciej Wołczyk, Bla.zej Osi'nski, and Peter Ondruska.
\newblock Urban driver: Learning to drive from real-world demonstrations using policy gradients.
\newblock In \emph{Conference on Robot Learning}, 2021.

\bibitem[Scheel et~al.(2022)Scheel, Bergamini, Wolczyk, Osi{\'n}ski, and Ondruska]{scheel2022urban}
Oliver Scheel, Luca Bergamini, Maciej Wolczyk, B{\l}a{\.z}ej Osi{\'n}ski, and Peter Ondruska.
\newblock Urban driver: Learning to drive from real-world demonstrations using policy gradients.
\newblock In \emph{Conference on Robot Learning}, pages 718--728. PMLR, 2022.

\bibitem[Seff et~al.(2023)Seff, Cera, Chen, Ng, Zhou, Nayakanti, Refaat, Al-Rfou, and Sapp]{seff2023motionlm}
Ari Seff, Brian Cera, Dian Chen, Mason Ng, Aurick Zhou, Nigamaa Nayakanti, Khaled~S Refaat, Rami Al-Rfou, and Benjamin Sapp.
\newblock Motionlm: Multi-agent motion forecasting as language modeling.
\newblock In \emph{Proceedings of the IEEE/CVF International Conference on Computer Vision}, pages 8579--8590, 2023.

\bibitem[Sha et~al.(2023)Sha, Mu, Jiang, Chen, Xu, Luo, Li, Tomizuka, Zhan, and Ding]{sha2023languagempc}
Hao Sha, Yao Mu, Yuxuan Jiang, Li Chen, Chenfeng Xu, Ping Luo, Shengbo~Eben Li, Masayoshi Tomizuka, Wei Zhan, and Mingyu Ding.
\newblock Languagempc: Large language models as decision makers for autonomous driving.
\newblock \emph{arXiv preprint arXiv:2310.03026}, 2023.

\bibitem[Shafiullah et~al.(2022)Shafiullah, Cui, Altanzaya, and Pinto]{shafiullah2022behavior}
Nur Muhammad~Mahi Shafiullah, Zichen~Jeff Cui, Ariuntuya Altanzaya, and Lerrel Pinto.
\newblock Behavior transformers: Cloning $k$ modes with one stone, 2022.

\bibitem[Sohl-Dickstein et~al.(2015)Sohl-Dickstein, Weiss, Maheswaranathan, and Ganguli]{SohlDickstein2015DeepUL}
Jascha~Narain Sohl-Dickstein, Eric~A. Weiss, Niru Maheswaranathan, and Surya Ganguli.
\newblock Deep unsupervised learning using nonequilibrium thermodynamics.
\newblock \emph{ArXiv}, abs/1503.03585, 2015.

\bibitem[Song et~al.(2020)Song, Meng, and Ermon]{song2020denoising}
Jiaming Song, Chenlin Meng, and Stefano Ermon.
\newblock Denoising diffusion implicit models.
\newblock \emph{arXiv preprint arXiv:2010.02502}, 2020.

\bibitem[Su et~al.(2021)Su, Lu, Pan, Murtadha, Wen, and Liu]{su2021roformer}
Jianlin Su, Yu Lu, Shengfeng Pan, Ahmed Murtadha, Bo Wen, and Yunfeng Liu.
\newblock Roformer: Enhanced transformer with rotary position embedding.
\newblock \emph{arXiv preprint arXiv:2104.09864}, 2021.

\bibitem[Surís et~al.(2023)Surís, Menon, and Vondrick]{surís2023vipergpt}
Dídac Surís, Sachit Menon, and Carl Vondrick.
\newblock Vipergpt: Visual inference via python execution for reasoning, 2023.

\bibitem[Tang and Salakhutdinov(2019)]{multiplefutures}
Yichuan~Charlie Tang and Ruslan Salakhutdinov.
\newblock Multiple futures prediction.
\newblock \emph{CoRR}, abs/1911.00997, 2019.

\bibitem[Urain et~al.(2023)Urain, Funk, Chalvatzaki, and Peters]{urain2022se}
Julen Urain, Niklas Funk, Georgia Chalvatzaki, and Jan Peters.
\newblock Se (3)-diffusionfields: Learning cost functions for joint grasp and motion optimization through diffusion.
\newblock \emph{ICRA}, 2023.

\bibitem[Wang et~al.(2022)Wang, Hunt, and Zhou]{wang2022diffusion}
Zhendong Wang, Jonathan~J Hunt, and Mingyuan Zhou.
\newblock Diffusion policies as an expressive policy class for offline reinforcement learning.
\newblock \emph{arXiv preprint arXiv:2208.06193}, 2022.

\bibitem[Wei et~al.(2022)Wei, Wang, Schuurmans, Bosma, Chi, Le, and Zhou]{DBLP:journals/corr/abs-2201-11903}
Jason Wei, Xuezhi Wang, Dale Schuurmans, Maarten Bosma, Ed~H. Chi, Quoc Le, and Denny Zhou.
\newblock Chain of thought prompting elicits reasoning in large language models.
\newblock \emph{CoRR}, abs/2201.11903, 2022.

\bibitem[Wen et~al.(2023)Wen, Yang, Fu, Wang, Cai, Li, Ma, Li, Xu, Shang, et~al.]{wen2023road}
Licheng Wen, Xuemeng Yang, Daocheng Fu, Xiaofeng Wang, Pinlong Cai, Xin Li, Tao Ma, Yingxuan Li, Linran Xu, Dengke Shang, et~al.
\newblock On the road with gpt-4v (ision): Early explorations of visual-language model on autonomous driving.
\newblock \emph{arXiv preprint arXiv:2311.05332}, 2023.

\bibitem[Williams et~al.(2015)Williams, Aldrich, and Theodorou]{williams2015model}
Grady Williams, Andrew Aldrich, and Evangelos Theodorou.
\newblock Model predictive path integral control using covariance variable importance sampling.
\newblock \emph{arXiv preprint arXiv:1509.01149}, 2015.

\bibitem[Xian et~al.()Xian, Gkanatsios, Gervet, Ke, and Fragkiadaki]{XianChainedDiffuserUT}
Zhou Xian, Nikolaos Gkanatsios, Th{\'e}ophile Gervet, Tsung-Wei Ke, and Katerina Fragkiadaki.
\newblock Chaineddiffuser: Unifying trajectory diffusion and keypose prediction for robotic manipulation.
\newblock In \emph{CoRL 2023}.

\bibitem[Xu et~al.(2019)Xu, Mart{\'{\i}}n{-}Mart{\'{\i}}n, Huang, Zhu, Savarese, and Fei{-}Fei]{DBLP:journals/corr/abs-1909-13072}
Danfei Xu, Roberto Mart{\'{\i}}n{-}Mart{\'{\i}}n, De{-}An Huang, Yuke Zhu, Silvio Savarese, and Li Fei{-}Fei.
\newblock Regression planning networks.
\newblock \emph{CoRR}, abs/1909.13072, 2019.

\bibitem[Xu et~al.(2023)Xu, Zhang, Xie, Zhao, Guo, Wong, Li, and Zhao]{xu2023drivegpt4}
Zhenhua Xu, Yujia Zhang, Enze Xie, Zhen Zhao, Yong Guo, Kenneth~KY Wong, Zhenguo Li, and Hengshuang Zhao.
\newblock Drivegpt4: Interpretable end-to-end autonomous driving via large language model.
\newblock \emph{arXiv preprint arXiv:2310.01412}, 2023.

\bibitem[Yang et~al.(2023)Yang, Du, Ghasemipour, Tompson, Schuurmans, and Abbeel]{Yang2023LearningIR}
Mengjiao Yang, Yilun Du, Kamyar Ghasemipour, Jonathan Tompson, Dale Schuurmans, and Pieter Abbeel.
\newblock Learning interactive real-world simulators.
\newblock \emph{ArXiv}, abs/2310.06114, 2023.

\bibitem[Yu et~al.(2023)Yu, Gileadi, Fu, Kirmani, Lee, Arenas, Chiang, Erez, Hasenclever, Humplik, et~al.]{yu2023language}
Wenhao Yu, Nimrod Gileadi, Chuyuan Fu, Sean Kirmani, Kuang-Huei Lee, Montse~Gonzalez Arenas, Hao-Tien~Lewis Chiang, Tom Erez, Leonard Hasenclever, Jan Humplik, et~al.
\newblock Language to rewards for robotic skill synthesis.
\newblock \emph{arXiv preprint arXiv:2306.08647}, 2023.

\bibitem[Zeng et~al.(2019)Zeng, Luo, Suo, Sadat, Yang, Casas, and Urtasun]{zeng2019end}
Wenyuan Zeng, Wenjie Luo, Simon Suo, Abbas Sadat, Bin Yang, Sergio Casas, and Raquel Urtasun.
\newblock End-to-end interpretable neural motion planner.
\newblock In \emph{Proceedings of the IEEE/CVF Conference on Computer Vision and Pattern Recognition}, pages 8660--8669, 2019.

\bibitem[Zhang et~al.(2021)Zhang, Liniger, Dai, Yu, and Van~Gool]{roach}
Zhejun Zhang, Alexander Liniger, Dengxin Dai, Fisher Yu, and Luc Van~Gool.
\newblock End-to-end urban driving by imitating a reinforcement learning coach.
\newblock In \emph{Proceedings of the IEEE/CVF International Conference on Computer Vision (ICCV)}, 2021.

\bibitem[Zhong et~al.(2022{\natexlab{a}})Zhong, Rempe, Xu, Chen, Veer, Che, Ray, and Pavone]{Zhong2022GuidedCD}
Ziyuan Zhong, Davis Rempe, Danfei Xu, Yuxiao Chen, Sushant Veer, Tong Che, Baishakhi Ray, and Marco Pavone.
\newblock Guided conditional diffusion for controllable traffic simulation.
\newblock \emph{ArXiv}, abs/2210.17366, 2022{\natexlab{a}}.

\bibitem[Zhong et~al.(2022{\natexlab{b}})Zhong, Rempe, Xu, Chen, Veer, Che, Ray, and Pavone]{controllabletraffic}
Ziyuan Zhong, Davis Rempe, Danfei Xu, Yuxiao Chen, Sushant Veer, Tong Che, Baishakhi Ray, and Marco Pavone.
\newblock Guided conditional diffusion for controllable traffic simulation.
\newblock \emph{ArXiv}, abs/2210.17366, 2022{\natexlab{b}}.

\bibitem[Zhu et~al.(2020)Zhu, Tremblay, Birchfield, and Zhu]{DBLP:journals/corr/abs-2012-07277}
Yifeng Zhu, Jonathan Tremblay, Stan Birchfield, and Yuke Zhu.
\newblock Hierarchical planning for long-horizon manipulation with geometric and symbolic scene graphs.
\newblock \emph{CoRR}, abs/2012.07277, 2020.

\bibitem[Ziegler et~al.(2014)Ziegler, Bender, Dang, and Stiller]{ziegler2014trajectory}
Julius Ziegler, Philipp Bender, Thao Dang, and Christoph Stiller.
\newblock Trajectory planning for bertha—a local, continuous method.
\newblock In \emph{2014 IEEE intelligent vehicles symposium proceedings}, pages 450--457. IEEE, 2014.

\end{thebibliography}
